# FogROS G: Enabling Secure, Connected and Mobile Fog Robotics with Global Addressability

Kaiyuan Chen[1], Jiachen Yuan[1], Nikhil Jha[1], Jeffrey Ichnowski[1],
John Kubiatowicz[1], and Ken Goldberg[1,2]

*Abstract*— Fog Robotics renders networked robots with greater mobility, on-demand compute capabilities and better energy efficiency by offloading heavy robotics workloads to nearby Edge and distant Cloud data centers. However, as the de-facto standard for implementing fog robotics applications, Robot Operating System (ROS) and its successor ROS2 fail to provide fog robots with a mobile-friendly and secure communication infrastructure.

In this work, we present FogROS G, a secure routing framework that connects robotics software components from different physical locations, networks, Data Distribution Service (DDS) and ROS distributions. FogROS G indexes networked robots with globally unique 256-bit names that remains constant even if the robot roams between multiple administrative network domains. FogROS G leverages Global Data Plane, a global and secure peer-to-peer routing infrastructure between the names, guaranteeing that only authenticated party can send to or receive from the robot. FogROS G adopts a proxy-based design that connect nodes from ROS1 and ROS2 with mainstream DDS vendors; this can be done without any changes to the application code. The code is publicly available at **https://github.com/KeplerC/gdp-for-ros**.

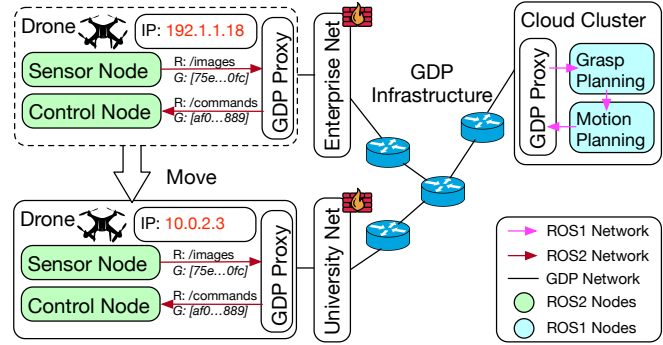

Fig. 1: FogROS G connects fog robotics drone with cloud compute resources. When the drone flies from an enterprise network to a university network, FogROS G persists its ROS connection over topic `/images` and `/commands`, despite (1) its IP address has changed and (2) both IP addresses are behind firewall and not publicly accessible (2) the nodes on the cloud are implemented in ROS1 but drone uses ROS2. This is accomplished by assigning every ROS topic a globally-unique and location-independent name(`[75e...0fc]` and `[af0...889]`) and by securely routing the messages through GDP infrastructure.

## I. INTRODUCTION

As robot onboard resources fail to keep up with heavier robotics computations, leveraging external computational resources on demand becomes more desirable. Fog robotics [1], [2] has been proposed for the continuum of Cloud Computing that offers immense on-demand computational resources and Edge Computing that exploits the resources close to the robot with lower access latency [3], [4]. In this paper, we propose FogROS G, a framework that enables secure and location-independent routing for Robot Operating System (ROS)[1]. It allows networked robots to roam freely with no requirement of a static network address (i.e. IP), network administrative domain or physical location. With minimal effort, robotic application developers can publish their compute services globally and only authenticated robots can use the services; they can also establish secure communication channel with robots even if the robots do not have a publicly accessible network address.

ROS is the de-facto standard for implementing fog robotics applications. It modularizes the robotics application into *nodes*, and connects the nodes into a graph. Nodes communicate with each other through a publish-subscribe (pub/sub) system that node publishes messages with respect to *topics*, and subscribes to the topics published by other nodes.

Although it presents a convenient interface to robotics application developers, the current ROS 2 implementation cannot meet the following requirements of the fog robotics applications:

*a) Edge-Cloud Continuum:* ROS computational graph may span across multiple network administrative domains with the recent advances of the edge computing and multi-cloud computing [5], [6], [7], [8]. For example, in FogROS 2[9], a ROS node on the robot can connect to both Google Cloud Platform (GCP) and Amazon Web Services (AWS). However, some of the IP addresses are visible only within the same network domain, and external accesses are blocked by firewall. ROS is unable to discover the nodes out of the domain and those without a publicly accessible address.

*b) Mobility:* ROS uses static IP address to discover and peer with ROS nodes on other machines. However, because an IP address is usually bound to a physical location. The mobile robots, such as Unmanned Aerial Vehicle, cannot have a static IP address. Existing approaches such as mobile IP, are known to have security and reliability issues [10].

*c) Security:* ROS is notorious for the lack of the security mechanisms [11], [12]. For example, due to the lack of authentication scheme in the ROS implementation, anyone can join an existing ROS network, publish to and subscribe from the robots. Attackers can freely listen to the existing

[1]Department of Electrical Engineering and Computer Science
[2]Department of Industrial Engineering and Operations Research
[1,2]University of California, Berkeley, CA, USA
{kych, jcyuan_johnson, nikhiljha, jeffi, kubitron, goldberg}@berkeley.edu

[1]For the rest of the paper, we use ROS for both ROS1 and ROS2.

communication and remotely control the robot [13] .

*d) Compatibility:* ROS2 fails in both backward-compatibility with ROS1 and unification with different DDS vendors [14], [15]. While most of the existing research prototypes are implemented with ROS1, ROS2 nodes cannot directly communicate with running ROS1 nodes. In addition, network configuration with ROS2 is DDS vendor-specific: different vendors use different configuration languages and offer different features. Most of the existing multiple machine implementations with ROS2 only work with one of the DDS, such as FogROS [16] and Zenoh [17].

Fog Robotics challenges the traditional wisdom to retrains robots behind firewalls, which hinders the mobility of the robots and limits the ability to leverage external computational resources. Thus, FogROS G argues a paradigm shift to make ROS nodes **location-independent**. Every ROS node is indexed with a globally visible and unique name, which is independent from the location of the robot and the network that it connects to. The fog robots can roam freely: as long as there is a globally connected path in FogROS G, they can stay connected with their edge or cloud services. We adopt a **security-first** design that only the authenticated party can connect to the node and establish a secure communication with it.

In FogROS G, every ROS topic is indexed with 256-bit name derived from the meta information (such as the name, author, functionality) of node. The attacker is unable to guess the name, because there are $2^{256}$ possible names and modifying one bit of the meta information changes the 256-bit name entirely (i.e. avalanche effect). The names are mapped to Global Data Plane (GDP) [18], a federated peer-to-peer network that can securely route the messages across different network administrative domains. We design a GDP-ROS proxy that converts between ROS messages and GDP packets, and cryptographically harden the packets with confidentiality, integrity and authenticity. The proxy-based designs also allow to connect ROS nodes with different network middleware vendors and ROS distributions.

In this paper, we implement GDP routing infrastructure in Rust, a memory-safe and type-safe language. The switches are implemented on kernel-bypass DPDK for minimal routing overhead. We implement a proxy between GDP and ROS similar to FogROS [16], but the packets are sent in layer 2 with GDP headers.

FogROS G aims the following design goals:
1) *Security* FogROS G assumes the attackers may tap into any network links between robots and cloud servers, and they can eavesdrop and tamper the ongoing communication. FogROS G prevents them from learning any of the plaintext data (such as topic name, ROS messages), and detect any unauthenticated messages sent to the ROS nodes.
2) *Location Independent* A node with FogROS G stays connected with its ROS graph as long as it can connect to one of the other nodes enabled with FogROS G (even if they do not belong to the same ROS graph). Other ROS nodes are able to connect to the node with the same interface and 256-bit name.
3) *User-Friendly* FogROS G does not require **any** modifications to the original ROS 2 application code. Application developers only need to specify the nodes intended to be global in a separate configuration file. This is helpful for separating security policy with actual implementation.
4) *Compatibility* FogROS G does not rely on any of the DDS-specific configuration, and can be ported to both ROS1 and ROS2. It also allows ROS nodes with different DDS vendors and ROS distributions connected with the same FogROS G network.

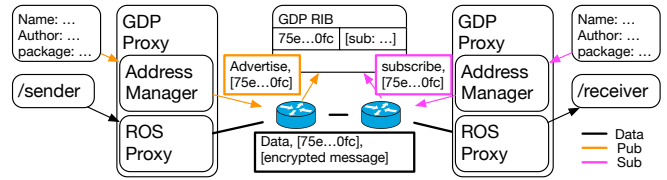

Fig. 2: An overview of the proposed FogROS G architecture. (**Orange**) shows the procedure of generating and publishing a 256-bit name [75e...0fc] globally to GDP Routing Information Base (RIB). (**Magenta**) shows the subscriber can derive the same name [75e...0fc] with the meta-information of the topic and send the subscription query. (**Black**) The GDP infrastructure populates the switches with associated routing information, and forward the actual encrypted data.

## II. DESIGN

Figure 2 shows an overview of the proposed architecture. FogROS G has three main components: (1) an address manager that converts and registers ROS node metadata to 256 bit globally unique name (2) GDP routing infrastructure (3) a proxy module that converts between ROS messages and GDP packets.

To make a ROS topic globally visible, the user specifies the meta-information (such as author, functionality) about the topics. The meta-information is converted to a 256-bit globally unique name by SHA-256 hash, and publishes the topic information to GDP Routing information Base (RIB). FogROS G adopts a proxy-based approach that serializes the ROS messages with vendor-independent encoding and converts them to GDP packets. The GDP packets are cryptographically hardened with confidentiality and integrity, and they are routed to the target FogROS G proxy. After the GDP packets are decrypted and verified their authenticity. they are converted back to ROS messages and sent to the destination ROS node.

### A. Global Addressable ROS

FogROS G uses ROS topic as the minimal granularity of the global name, because a robot can run many ROS nodes, and a ROS node can host many ROS topics. ROS topic exposes a standardized interface with fixed message format. It can be easily extended to other ROS features, such as services. User can restrict the exposure of ROS nodes by allowing only part of the interfaces to public. Partitioning public and private interfaces also enhance the privacy and isolation of the ROS nodes. If one uses ROS Master in ROS1 or VPN in FogROS[16], all nodes have to join the same pub/sub network. A node may accidentally subscribe to a topic with the same name on another machine, which greatly introduces unwanted message exchanges and additional communication overhead.

The global name is designed to be unique, static and location independent, with the goal that anyone can derive and identify the topic even if he or she does not know the 256-bit name. The name is derived from the metadata of the ROS topic, such as the ROS node's name, author, maintainer, interface, description etc. It can also contain a human-memorable random number [19] to prevent multiple identical ROS nodes running at the same time. Optionally, one can include the public portion of the security credential to facilitate the verification of the name advertisement.

The metadata is serialized and converted into 256-bit string using SHA-256 hash. There are several reasons why hashed string is suitable for using as the globally unique name: (1) **deterministic** The hash is deterministic so that every party holding the same metadata can derive the same hash value and thus the same global name for the ROS node (2) **Oneway** SHA-256 is a one way function that the attacker cannot reverse the original meta-data from the 256 bit name and thus does not leak the original metadata. (3) **Avalanche effect** A small change to the original metadata leads to new hash value that appear unrelated to the original hash value (4) **Large name space** There are $2^{256}$ possible names and it is proved to be computationally infeasible to find two messages with the same hash.

### B. GDP-ROS proxy

The GDP-ROS proxy converts between ROS messages and GDP packets. After user initializes the proxy with the metadata of the topics, the proxy subscribes to the events of the topic. For example, if there is a new local publisher ROS node joins the network, it publishes the topic globally. Using proxy-based design has several advantages:

*a) Vendor-Agnostic:* Existing efforts[9] on connecting multiple ROS nodes across multiple machines require vendor-specific configurations, and every vendor requires different configuration format [20], [21]. As a result, most of the existing solutions only work for one DDS (for example, FogROS 2 and Zenoh [17] only support CycloneDDS [14]). However, FogROS G's proxy presents an abstraction of a *virtual* node that is agnostic to the heterogeneous implementations of the ROS nodes. Due to the same reason, it *works for both ROS1 and ROS2*.

*b) Efficient and Adaptive:* The proxy does not introduce additional overhead if all the ROS nodes are on the same machine, as they still use the original ROS to manage the graph and inter-node communication. It does not exposes the topics to public unless specified by the user.

*c) User-Friendly:* The proxy presents a virtual node abstraction to other nodes so that other nodes can communicate with the node in the same way as they communicate with a local ROS node. As a result, it allows ROS node to be transparent about the underlying routing infrastructure and users can interact with remote nodes without modifying the original application code.

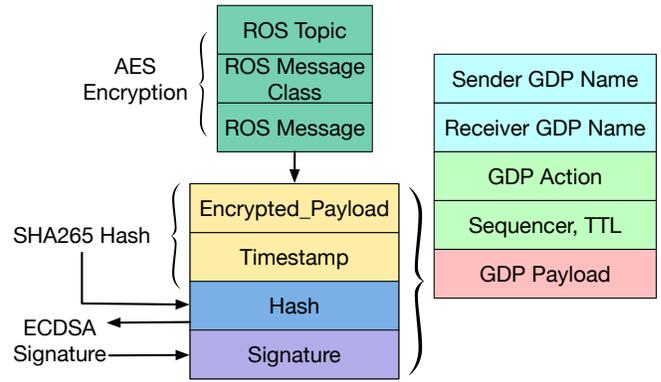

Fig. 3: An illustration of the fields of GDP packets. ROS messages are serialized into JSON formatted string. The string is encrypted, hashed and signed for confidentiality and integrity.

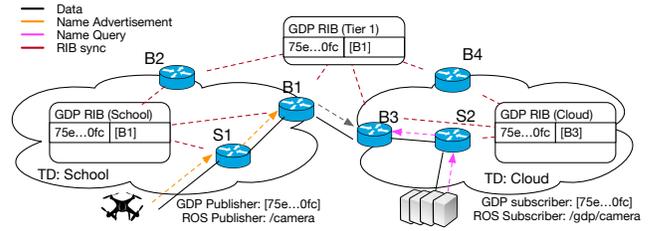

Fig. 4: An illustration of the routing architecture of the GDP network with two Trust Domains (School and Cloud). There is a higher tier RIB that connect both domains and facilitate the query of the name.

### C. FogROS G Data Security

All FogROS G communication is exchanged in GDP packets that guarantee the confidentiality, integrity and authenticity of the ROS messages. Figure 3 shows how a GDP packet is generated with ROS. The ROS message is serialized into a string that contains ROS message type, topic and the original message. The string is encrypted with AES encryption [22] so that only the parties with the same symmetric key is able to decrypt the the ciphertext to the plaintext. The encrypted data is hashed with SHA-256 [23] to protect the integrity of the data, that modifying one bit of the payload changes the resultant hash entirely. The hash is signed with the private portion of the writer key with Elliptic Curve Digital Signature Algorithm(ECDSA) [24], so that anyone holding the public portion of the writer key is able to verify that the content is produced by the authentic writer. The resultant payload is used as the GDP payload sent through the GDP routing infrastructure.

### D. Location-Independent Routing with GDP

Global Data Plane (GDP) is a peer-to-peer routing infrastructure that enables secure routing on federated network infrastructure that there is no central authority for assigning names. GDP assigns all the entities (network hardware, data, message senders and receivers) with *flat* and *location-independent* names. These names do not involve any physical identifiers (such as IP addresses), allowing the resources to be placed, moved and replicated to different locations. The location-independent name also allows the abstraction that one can directly communicate with services, data instead of an end host.

GDP enables flat and location-independent routing with a routing fabric of switches and Routing Information Bases (RIB). GDP organizes the routing fabric as a number of network administrative domains named Trust Domains (TDs) and restrict the routing only through trusted TDs. TD guarantees every domain can independently maintain verifiable routing state. GDP use a hierarchical structure of TDs to ensure the routing scalability and locality.

Figure 4 illustrates a GDP network with two TDs. The robot drone publishes a GDP name [75e...0fc] to its local RIB, and the routing information is flooded to higher tier RIB through the border switches B1 and B2. The RIBs verify the advertisement message by validating the hash generation and the cryptographic credentials of the name advertiser. After verification, RIB updates the 256-bit name and its corresponding switch to its in-memory key-value store. The remote subscriber from the cloud TD sends a name query, and the name query is propagated to its local RIB and the higher tier RIB. The GDP RIB (Tier 1) responds the query with B1's routing information, and B3 subscribes to the topic by peering with B1. Finally ROS messages are published through the switches along the path with shortest path routing algorithm to the cloud subscriber.

## III. PRELIMINARY EVALUATION

### A. Implementation

We implement GDP routing infrastructure with 23,378 Lines of Code in Rust. Rust is a type-safe language with strong memory and type safety at compile time. We use Capsule, Comcast's implementation of NetBricks [25], to implement the switching and RIB. Capsule wraps around Intel's Data Plane Development Kit (DPDK) to bypass the kernel and provide fast userspace packet processing ability.

We implement the proxy with 2,000 LoC in Python. The proxy interfaces the corresponding ROS version and serializes the ROS messages to JSON format. The proxy interfaces the GDP infrastructure by sending raw bytecode packets through `sendp` through layer 2 (versus regular layer 3 `send` used by DDS). Because Layer 2 packets are required to have fixed length of payload (typically 1500 bytes), the proxy also implements packet re-ordering and assembly mechanism.

### B. Evaluation Setup

We use five Intel NUC edge servers with an Intel® Pentium® Silver J5005 CPU @ 1.50 GHz with 2 cores enabled to emulate the robot and 4 cores enabled to emulate the edge servers. They are connected in a ring that one server is only connected to two other neighbor servers. We conduct preliminary evaluation on the networking performance with ROS 2 and CycloneDDS with three real robotics applications.

FogROS G is evaluated with the following ROS applications: (1) Visual SLAM Simultaneous Localization and Mapping (vSLAM) with ORB-SLAM2 [26], and we use

|          | Robot | ROS2    |       | FogROS G |       |
|----------|-------|---------|-------|----------|-------|
| Scenario | Only  | Network | Total | Network  | Total |
| vSLAM         | 0.68  | 0.010 | 1.4 | 0.024 | 1.4 |
| MotionPlanning| 161.8 | 0.008 | 4.3 | 0.023 | 4.3 |
| GraspPlanning | 14.0  | 0.065 | 0.6 | 0.089 | 0.6 |

TABLE I: **Network and end-to-end total latency benchmark of FogROS G on edge.** We measure the network and total time **in seconds** using unsecured ROS2 with CycloneDDS on the edge, FogROS G with the cloud and edge servers. Total time is the sum of network and compute time. We assume both cloud server and robot run behind firewall similar to Figure 4.

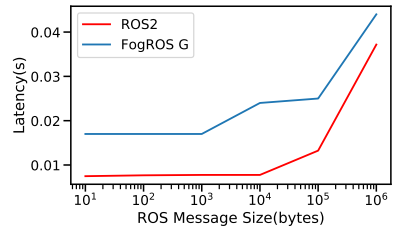

Fig. 5: **Network Latency Comparison between Off-the-shelf ROS2 and FogROS G**. The experiment is conducted through three intermediate routing nodes.

`fr2/loop` from TUM Dataset[11]; (2) Grasp Planning with Dex-Net [27], and (3) Motion Planning with Motion Planning Templates (MPT) [28], and we use Home dataset from Open Motion Planning Library OMPL[29]. We refer readers to FogROS1 [16] and FogROS2 [9] for more details.

Figure 5 compares the network latency between off-the-shelf ROS2 and FogROS G. FogROS G introduces an additional 0.014 second overhead on average compared to the off-the-shelf ROS2 with CycloneDDS. The additional latency is caused by the proxy and the cost of cryptographic operations. Due to our current implementation limitations, the numbers for FogROS G show the one-way latency instead of the end-to-end latency. The actual latency will be slightly larger and they will be reflected in the final published version.

The key insight from Table I is that FogROS G introduces very little overhead compared to the the original unsecured ROS2 - despite FogROS G is safer and more connected. The overhead is negligible to the actual compute intensive application.

## IV. DISCUSSION AND FUTURE WORK

In this paper, we enable secure and location independent fog robotics applications by giving every ROS node a globally identifiable name and using GDP to securely route to the name. Besides completing the evaluation section and deploying our design with real robots, we deem the following research directions as the future work of this project: (1) *ROS message persistence and replay*. We hope to leverage the persistence aspect of the GDP routing. As the ROS messages can be securely routed to persistent storage server and replayed later. (2) *Service-centric Anycast Routing*. In this paper, we explored indexing a ROS node with globally unique name. However, we can name *services* (such as grasp planning as a service) with the same 256-bit name. The robot can query in an anycast style to reach out to the service closest to the robot.